\documentclass{article}
\usepackage{spconf,amsmath,graphicx}
\usepackage{xcolor}
\usepackage{flushend}
\usepackage{url}

\title{Unsupervised pre-trained, texture aware and lightweight model for deep learning based iris recognition under limited annotated data}
\name{Manashi Chakraborty$^{\dagger \ddagger}$ \thanks{$^ \dagger$All correspondence to : manashi.chakraborty@iitkgp.ac.in} \thanks{$^\ddagger$Denotes equal contribution}\qquad Mayukh Roy$^{\ddagger}$  \qquad Prabir Kumar Biswas \qquad Pabitra Mita}
\address{Indian Institute of Technology, Kharagpur, India}

\begin{document}
%
\maketitle
\begin{abstract}
In this paper, we present a texture aware lightweight deep learning framework for iris recognition. Our contributions are primarily three fold. Firstly, to address the dearth of labelled iris data, we propose a reconstruction loss guided unsupervised pre-training stage followed by supervised refinement. This drives the network weights to focus on discriminative iris texture patterns. Next, we propose several texture aware improvisations inside a Convolution Neural Net to better leverage iris textures. Finally, we show that our systematic training and architectural choices enable us to design an efficient framework with upto 100$\times$ fewer parameters than contemporary deep learning baselines yet achieve better recognition performance for within and cross dataset evaluations.

\end{abstract}
\begin{keywords}
Iris Recognition, Deep Learning, CNN, Texture, Lightweight
\end{keywords}
\section{Introduction}
\label{sec:intro}

Iris biometrics, over the last few years have shown immense potential as an infallible biometric recognition system \cite{1262028, daugman1993high, de2001iris, wildes1996machine, wildes1997iris}. Iris textures are highly subject discriminative \cite{daugman1993high} and being an internal organ of the eye, it is resilient to environmental perturbations and is also immutable over time. 


    \par The initial works on iris recognition focused on designing traditional hand engineered features \cite{1262028, masek2003recognition, monro2007dct, ma2003personal}. Recent success over a variety of vision applications on natural images \cite{girshick2015fast, krizhevsky2012imagenet} showcases the unprecedented advantage of deep Convolution Neural Networks (CNNs) over hand-crafted features. Inspired by the success of CNNs, iris biometric community also started exploring the prowess of deep learning. An appreciable gain in performance \cite{minaee2016experimental,gangwar2016deepirisnet, nguyen2017iris} is observed compared to traditional methods. However, some intrinsic issues such as absence of large annotated datasets, explicit processing of texture information and lightweight architecture design have hardly been addressed. In this paper, we address the above concerns with several systematic modifications over conventional CNN training pipelines and architectural choices.\\
\textbf{Handling Absence of Large Dataset:}
CNNs are data greedy and usually require millions of annotated data for fruitful training. This is not an issue for natural images where datasets such as Imagenet \cite{deng2009imagenet}, MS-COCO \cite{lin2014microsoft} contain large volumes of annotated data. However, for iris biometrics, the sizes of the datasets are usually limited to few thousands. Thus, this short-coming still remains an open challenge for deep learning based iris biometric researchers. In this paper, we address this problem with a two-stage training strategy. In the first stage, we pre-train a parameterized feature encoder, $E_{\theta}(\cdot)$, to capture iris texture signatures in an unsupervised training framework. In the second stage, $E_{\theta}(\cdot)$  acts as a feature extractor and is further refined along with a classification head, $C_{\psi}(\cdot)$. We show that the combined training framework provides significant boost in performance compared to single stage training. Further, visualization with  Layer Wise Relevance Propagation \cite{bach2015pixel} shows that as opposed to single-stage training, our proposed stage-wise training drives the network weights to focus more on the iris textures. This further motivated us in designing systematic texture attentive architectural choices as mentioned below. \\
\textbf{Energy Aware Pooling:} 
Non-parametric spatial sub-sampling (usually realised as Max-pooling) in conventional deep networks is a crucial and essential component fairly used to retain the maximum response of a specified window. In this paper, we show that on a texture-rich iris \cite{daugman1993high} dataset, sub-sampling using Energy Aware Pooling (\textit{EAP}) is a better alternative to  $max(\cdot)$ operation.\\
\textbf{Texture Energy Layer:} Usually in deep networks, it is a common practise to have several fully-connected layers at the end to amalgamate global structure information. However, iris images are mainly rich in local textures. Toward this, we propose to use Texture Energy Layer (\textit{TEL}) to specifically capture energy of the last convolutional filter bank responses. Such energy based features have been traditionally used for texture classification \cite{han2007rotation,unser1995texture,idrissa2002texture}.\\
\textbf{Light-weight Model for Inference:} 
The systematic design strategies enable us to operate with much shallower architecture yet achieve better performance than the deeper baselines.  
Additionally, \textit{TEL} layer  obviates the requirement of computationally heavy penultimate fully-connected layer of our proposed base architecture. As a consequence, our model has significantly less parameter counts. This is particularly important since iris biometrics is gradually becoming an integral component of many handheld mobile devices.
\par Our above proposed architectural choices consistently outperforms traditional as well as recent deep nets by a noteworthy margin. Even in scenarios where target dataset is different from training data, our proposed model generalises with better performance without the need of even fine-tuning on the target data.


\begin{figure}[!t]
    \centering
    \includegraphics[width=8.5cm, height=3.5cm]{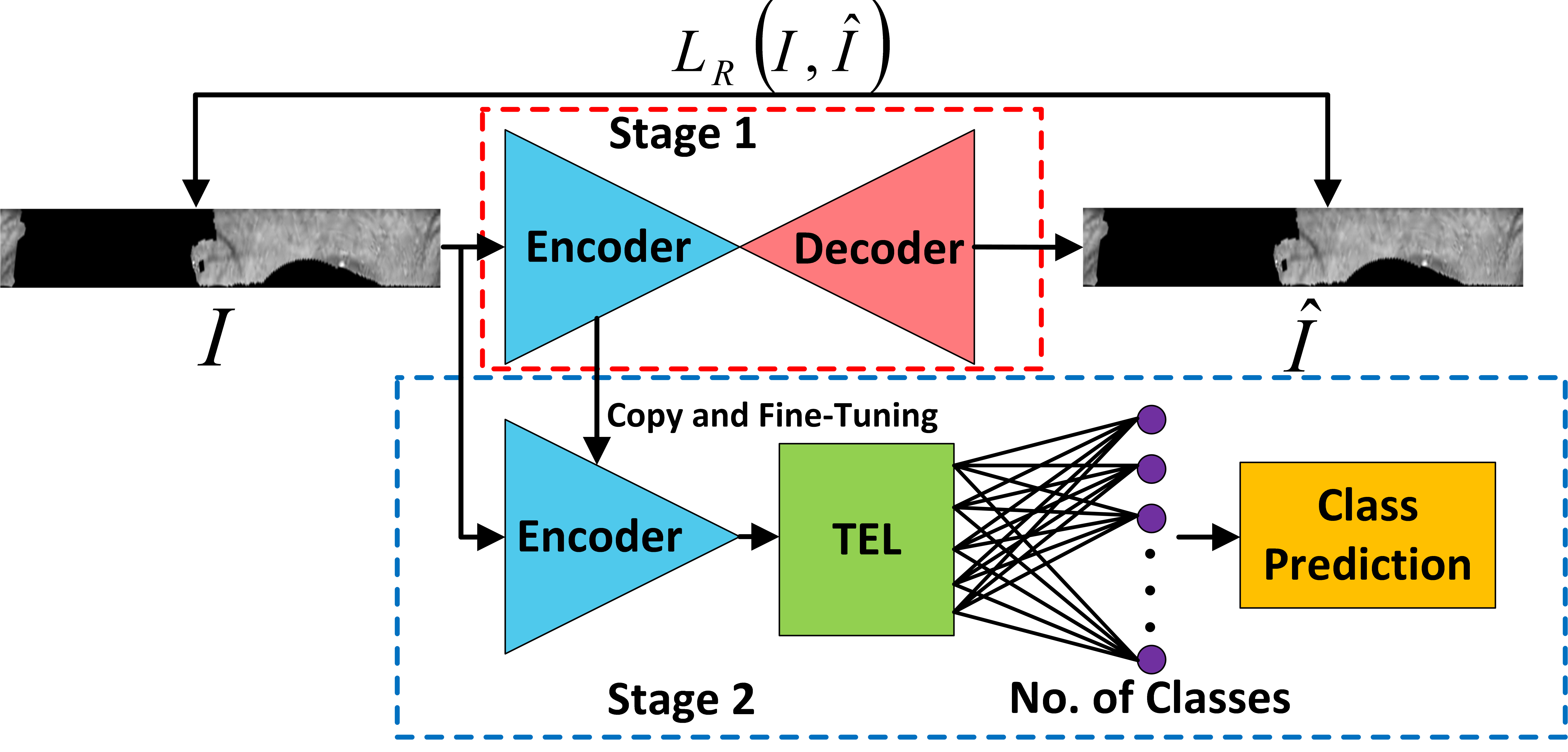}
    \vspace{-3mm}
    \caption{\scriptsize Stagewise training framework of proposed framework of $CombNet_{E_{\theta}}^{EAP+TEL}$}
    \label{fig:framework}
    \vspace{-5mm}
\end{figure}

\vspace{-4mm}
\section{Related Work}
\label{sec:literature}
\vspace{-2mm}
Initial attempts of iris recognition were primarily inclined towards traditional techniques of extracting features from various filter bank responses. Daugman \cite{1262028} extracted representative iris features from responses of 2-D Gabor filters. Masek \textit{et al.} extracted response from 1D Log Gabor filters \cite{masek2003recognition}. Ma \textit{et al.} \cite{ma2003personal} proposed a bank of circularly symmetric sinusoidal modulated Gaussian filters banks to capture the discriminative iris textures. Wildes \textit{et al.}~\cite{wildes1997iris} extracted discriminative iris textures from multi-scale Laplacian of Gaussian (LOG). Monro \textit{et al.} used features from Discrete Cosine Transform (DCT) \cite{monro2007dct}. To summarize, the earlier works mainly focused on handcrafted feature representation. Initial attempts \cite{minaee2016experimental,nguyen2017iris} of leveraging deep learning for iris recognition involved feature extraction using well known pre-trained (for ImageNet classification) neural networks followed by a supervised classification stage. 
Recently, Gangwar et al.~\cite{gangwar2016deepirisnet} proposed  DeepIrisNet, which is an to end-to-end trainable (from scratch) deep neural network and achieved appreciable boost over the  traditional methods.
\vspace{-4mm}
\section{Methodology}
\vspace{-2mm}
\label{sec:method}

\subsection{Network Architecture}
\label{sec:pagestyle}
\vspace{-2mm}
\subsubsection{Stagewise Training} 
\label{sec:training}
\vspace{-2mm}
\textbf{Stage-1:}
\label{sec:featinit}
In the first phase, we follow an unsupervised framework for pre-training a feature encoder, $E_{\theta}(\cdot)$ \footnote{subscript $\theta$ refers to set of trainable parameters} to capture texture signatures. For this, we train a convolutional auto-encoder with reconstruction loss, $L_R$. Specifically, given a normalised iris image, $I$ (an example of normalised iris image, $I$ is shown in Figure \ref{fig:framework}), we project it to a smaller resolution (by strided convolution and spatial sub-sampling) using the encoder and then decode it back to the original resolution with a decoder, $D_{\phi}(\cdot)$. Configurations of various layers of encoder, $E_{\theta}(\cdot)$ and  decoder, $D_{\phi}(\cdot)$ is shown in Table \ref{encoderDec}. $L_R$ is thus applied between original image, $I$ and reconstructed image, $\hat{I} =  D_{\phi}(E_{\theta}(I))$. In this paper, we have used the Structural Similarity (SSIM) metric $\in \{0, 1\}$ as a proxy for gauging the similarity between original and reconstructed image. So, we minimise the following:
\begin{equation}
    L_R = 1 - SSIM(I, D_{\phi}(E_{\theta}(I))).
\end{equation}

\textbf{Stage-2 \textit{CombNet}:} In the second stage, activations of $E_{\theta}(\cdot)$ is passed to the classification branch, $C_{\psi}(\cdot)$. Following the usual trend, the baseline $C_{\psi}(\cdot)$ consists of two fully connected layers followed by a softmax activation layer to output class probabilities. The combination of  $(E_{\theta}(\cdot),C_{\psi}(\cdot))$ is optimised using cross entropy loss. We term this combined architecture as $CombNet$. We define $CombNet_{E_{\theta}}$, as the combined model whose encoder, $E_{\theta}(\cdot)$ is pre-trained with reconstruction loss from Stage-1. $CombNet_{R}$ is the $CombNet$ model in which the encoder is randomly initialised (without any pre-training).
\begin{figure}[!t]
    \centering
    \includegraphics[width=8.5cm, height=1.0cm]{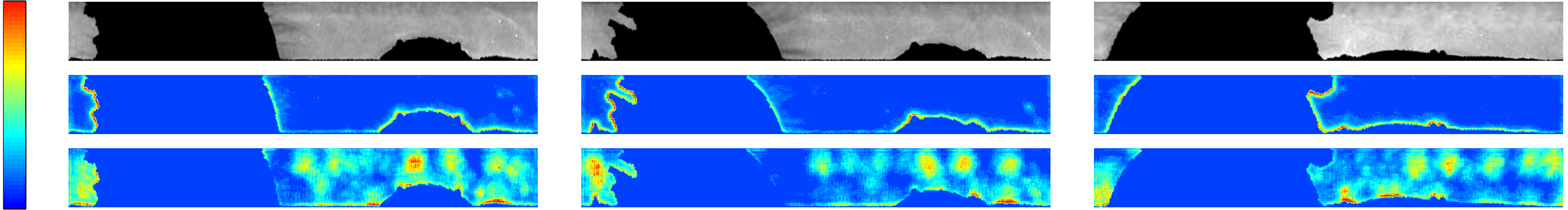}
    \vspace{-2mm}
    \caption{\scriptsize Relevance map (red is most important while blue is least) of three different iris corresponding to three classes of the CASIA.v4-Distance dataset. \textbf{Row 1:} Normalised iris image. \textbf{Row 2:} Relevance map of  $CombNet_{R}$ (randomly initialised encoder). \textbf{Row 3:} Relevance map of  $CombNet_{E_{\theta}}$ (initialised with pre-trained encoder).}
    \label{fig:lrp}
    \vspace{-6mm}
\end{figure}
\vspace{-3mm}
\subsubsection{Energy Aware Pooling \textit{(EAP)}:}
\vspace{-2mm}
This layer is proposed to retain the local texture energy during spatial sub-sampling in CNN. The de facto choice for sub-sampling in CNN is by Max-pool which is more appropriate to determine the presence/absence of a particular feature over the sampled window. For iris images which have local textural patterns, it is more prudent to retain the energy of the sub-sampled window. With this in mind, for a pooling kernel of receptive field $k \times k$, \textit{EAP} calculates the average of the $k^2$ pixels instead of finding the maximum as in Max-pool operation. Downsampling is achieved by operating this kernel with stride of 2 pixels. This way of retaining the energy while downsampling finds close analogy with energy of filter bank responses that has been traditionally used as discriminative feature for texture classification \cite{han2007rotation, idrissa2002texture}. We term the model with the proposed \textit{EAP} layer as $CombNet_{E_{\theta}}^{EAP}$.
\vspace{-3mm}
\subsubsection{Texture Energy Layer \textit{(TEL)}:} 
\vspace{-2mm}
This layer is designed to alleviate the need of penultimate fully connected layer of $CombNet_{E_{\theta}}^{EAP}$. This computationally heavy fully connected layer has entire image as its receptive field and thus looses local textures which are more important for iris recognition. Therefore, in this stage our present $CombNet_{E_{\theta}}^{EAP}$ is made more texture attentive by adding \textit{TEL} after the last convolution layer. In this layer we use spatial averaging kernels with spatial support equal to dimension of feature maps from previous layer. So, if input to \textit{TEL} layer is $H\times W \times C$, output from it is $1 \times 1\times C$ . These stacked average values closely corresponds to the energy of each activation maps of the previous layer. The output of \textit{TEL} is then finally passed to a single fully connected layer which is followed by softmax activation to get the final class probabilities. This combined texture attentive model having both \textit{EAP} and \textit{TEL} layers is termed as $CombNet_{E_{\theta}}^{EAP + TEL}$ which is shown in Figure \ref{fig:framework}. As \textit{TEL}  alleviates the need of penultimate fully connected layer, it helps in dramatically reducing the parameter count (46.72$\times$ cheaper) than our baseline having two fully connected layers as reported in Table \ref{ablation}. 
\vspace{-3mm}
\begin{table}[h!]
\centering
\vspace{-3mm}
\caption{\scriptsize Configurations of various layers of $E_{\theta}(\cdot)$ and  $D_{\phi}(\cdot)$}
\label{encoderDec}
\scalebox{0.58}
{
\begin{tabular}{lcccc}
\hline
\textbf{Type} & \textbf{Kernel} & \textbf{Stride} & \textbf{Padding} & \textbf{Output} \\
&&&&\textbf{Channels}\\\hline
\multicolumn{5}{c}{\textbf{Encoder}} \\ \hline
Conv & $5\times5$ & 1 & 2 & 32 \\
Batch Norm &  &  &  & 32 \\
Pooling & $2\times2$ & 2 & 0 & 32 \\
Conv & $3\times3$ & 1 & 1 & 64 \\
Batch Norm &  &  &  & 64 \\
Pooling & $2\times2$ & 2 & 0 & 64 \\
Conv & $3\times3$ & 1 & 1 & 128 \\
Batch Norm &  &  &  & 128 \\
Pooling & $2\times2$ & 2 & 0 & 128 \\
Conv & $3\times3$ & 1 & 1 & 256 \\
Batch Norm &  &  &  & 256 \\
Pooling & $2\times2$ & 2 & 0 & 256 \\
\hline
\multicolumn{5}{c}{\textbf{Decoder}} \\ \hline
Pixel Shuffle \cite{ledig2017photo} &  &  &  & 64 \\
Pixel Shuffle \cite{ledig2017photo} &  &  &  & 16 \\
Pixel Shuffle \cite{ledig2017photo} &  &  &  & 4 \\
Pixel Shuffle \cite{ledig2017photo} &  &  &  & 1 \\ \hline
\end{tabular}
}
\vspace{-3.5mm}
\end{table}
\subsection{Matching Framework}
\vspace{-2mm}
\label{sec:matching}
Representative iris signatures (1024-D) were extracted from the \textit{TEL} layer of $CombNet_{E_{\theta}}^{EAP + TEL}$. Two iris images are matched depending on the dissimilarity score obtained from the normalised euclidean distance between their respective iris signatures.
\vspace{-4mm}
\section{Experiments}
\vspace{-2mm}
\label{ssec:experiment}
\subsection{Comparing Methods}
\label{ssec:competing_methods}
\vspace{-2mm}
We compare our proposed framework with three traditional baselines: Daugman \cite{1262028}, Masek \cite{masek2003recognition} and Ma \textit{et al.} \cite{ma2003personal}. From deep learning paradigm, we compare against a pre-trained (on ImageNet) VGG-16 fined tuned on the iris dataset. This was one of the initial attempts of applying transfer learning with deep neural nets for iris data \cite{minaee2016experimental, nguyen2017iris}. We also compare against DeepIrisNet \cite{gangwar2016deepirisnet} which is a much deeper model having 8 convolution and 3 fully connected layer.
\vspace{-4mm}.
\subsection{Dataset Description}
\label{ssec:dataset}
\vspace{- 2mm}
We present our results on CASIA.v4-Distance \cite{casiav4dist} and CASIA.v4-Thousand \cite{casiav4dist}. Iris of left and right eye have disparate patterns \cite{daugman1993high} and are thus attributed to different classes i.e., number of classes is twice the number of subjects present in the dataset. 
\par The framework of \cite{zhao2015accurate} is used for iris segmentation and normalization. Normalised iris of three different subjects of CASIA.v4-Distance dataset is shown in Figure \ref{fig:lrp}. Spatial resolution of normalised iris images for all experiments is 512$\times$64 unless stated otherwise. For fair comparison, same segmentation and normalization protocol are followed for all experiments. We used the following two dataset configurations for performance evaluation.\\ 
\textbf{Within Dataset:} Here, \textit{`training+validation'} and test splits are selected from CASIA.v4-Distance dataset \cite{casiav4dist} having 142 subjects. Experiments were conducted on 4773 samples from 284 (left and right iris are considered as different classes) classes. Out of these 284 classes, \textit{`training+validation`} split comprises of 80$\%$ of the classes and the remaining disjoint 20$\%$ forms the test split used for reporting verification results (using matching framework of section \ref{sec:matching}).\\
    \textbf{Cross Dataset:}  In this setting, all the pre-trained models (trained on CASIA.v4-Distance) were directly used on CASIA.v4-Thousand dataset without any fine-tuning. This challenging configuration therefore evaluates the generalization capability of the different competing deep learning frameworks. CASIA.v4-Thousand has 2000 classes (left and right iris belong to different classes). We perform 5-fold testing. Each fold consists of $\frac{1}{5}^{th}$ of total classes. Average matching performance over the 5-folds is reported.
\par Following the matching framework of \cite{gangwar2016deepirisnet}, the test set for both the above configurations is divided into gallery (enrolled images) and probe (query) set. 50\% of the identities in probe set are imposters (identities not enrolled in the system) while the rest are genuine identities. 
\begin{table}[]
\scriptsize
\vspace{-3mm}
\centering
\caption{ \scriptsize Self ablation of various architectural choices.}
\label{ablation}
\begin{tabular}{|l|l|l|l|}
\hline
\multicolumn{1}{|c|}{\textbf{Model}} & \multicolumn{1}{c|}{\textbf{\begin{tabular}[c]{@{}c@{}}Classification Accuracy\\ 
(in \%)\end{tabular}}} &  \multicolumn{1}{c|}{\textbf{\begin{tabular}[c]{@{}c@{}}\#Params\\ \textbf{($10^6$)}\end{tabular}}} \\ \hline
$CombNet_{E_{\theta}}$  & 60.53 & 135.5 \\ 
$CombNet_{E_{\theta}}^{EAP}$  & 74.09 & 135.5 \\ 
\textbf{$CombNet_{E_{\theta}}^{EAP+TEL}$} & \textbf{92.61} & \textbf{2.9} \\ \hline
\end{tabular}
\vspace{-5mm}
\end{table}

\vspace{-3mm}
\subsection{Results}
\label{ssec:results}
\vspace{-2mm}

\textbf{Exp 1- Ablation study of various architectural  choices:} In this section, we perform self ablation of variants of architectural choices. We use classification accuracy on validation subset from the \textit{'training+validation'} split as a metric for model selection. Metrics are reported in Table \ref{ablation}.\\ 
\textit{a) Benefit of Stage-wise Training:} 
Classification accuracy of $CombNet_{E_{\theta}}$ is \textbf{60.53\%} while that of $CombNet_{R}$ is \textbf{53.11\%}. This clearly shows the benefit of pre-training the encoder part of $CombNet$ over random initialised encoder ($CombNet_{R}$).
Further, for reasoning the superiority of  $CombNet_{E_{\theta}}$ over $CombNet_{R}$, we study relevance map of a given iris image correctly classified by both the models. Relevance map gives an indication of which input pixels were important for classification. Fig \ref{fig:lrp} shows relevance (heat) map of both the aforementioned models from three different classes of CASIA.v4-Distance dataset. It is evident from figure that pre-training the encoder encourages $CombNet_{E_{\theta}}$ to focus more on the texture patterns as opposed to $CombNet_{R}$ which primarily concentrates on the overall shape cues obtained from the boundary (separating iris region from background) pixels. Instigated from this observation, we incorporate additional improvements on  $CombNet_{E_{\theta}}$ that further exploits the textural cues for better performance. \\ 
\textit{b) Benefit of \textit{EAP} and \textit{TEL} layers:} From Table \ref{ablation} we observe, as Max-Pool layer is replaced by \textit{EAP}, correspondingly classification accuracy increases from 60.53\% to 74.09\% . This bolsters our assumption that \textit{EAP} layer is more beneficial for sub-sampling than Max-Pool on texture-rich images. With replacement of the penultimate fully connected layer of $CombNet_{E_{\theta}}^{EAP}$ with \textit{TEL} layer, we see a further improvement of performance by our $CombNet_{E_{\theta}}^{EAP + TEL}$ model. \\
\textbf{Exp 2- Within and Cross dataset comparison of our preferred architecture with existing methods:} From Exp 1, it is clearly evident that $CombNet_{E_{\theta}}^{EAP + TEL}$ outperforms our other architectural choices. Therefore, in this phase comparison of our best architectural choice with existing traditional as well as deep learning models are presented. Performance is evaluated based on \textit{EER (Equal Error Rate)}, and \textit{AUC (Area Under the Curve)} of the Detection Error Tradeoff (DET) curve. We also report parameter counts of the competing deep nets which are metrics of computational complexity. 
\begin{table}[]
\scriptsize
\vspace{-3mm}
\centering
\caption{\scriptsize Comparison on CASIA.v4-Distance (within dataset configuration).}
\label{WithinDBcompare}
\begin{tabular}{|c|c|c|c|}
\hline
\textbf{Model} & \textbf{\begin{tabular}[c]{@{}c@{}}EER\\ (in \%)\end{tabular}} & \textbf{AUC} & \textbf{\begin{tabular}[c]{@{}c@{}}\#Params\\ (in $10^6$)\end{tabular}} \\ \hline
\multicolumn{4}{|c|}{\textit{\textbf{Traditional}}} \\ \hline
Masek \cite{masek2003recognition} & 5.70 & 0.030 & XXX \\
Li Ma \textit{et al.} \cite{ma2003personal} & 5.45 & 0.026 & XXX \\
Daugman \cite{1262028} & 5.20 & 0.015 & XXX \\ \hline
\multicolumn{4}{|c|}{\textit{\textbf{Deep Nets}}} \\ \hline
VGG-16 & 4.88 & 0.012 & 135.2 \\
DeepIrisNet \cite{gangwar2016deepirisnet} & 4.80 & 0.011 & 291.2 \\
\textbf{$CombNet_{E_{\theta}}^{EAP+TEL}$ (Proposed)} & \textbf{3.25} & \textbf{0.004} & \textbf{2.9} \\ \hline
\end{tabular}
\vspace{-3mm}
\end{table}

\par
\begin{figure}[!t]
    \vspace{-2mm}
    \centering

    \includegraphics[width=8.5cm, height=3.3cm]{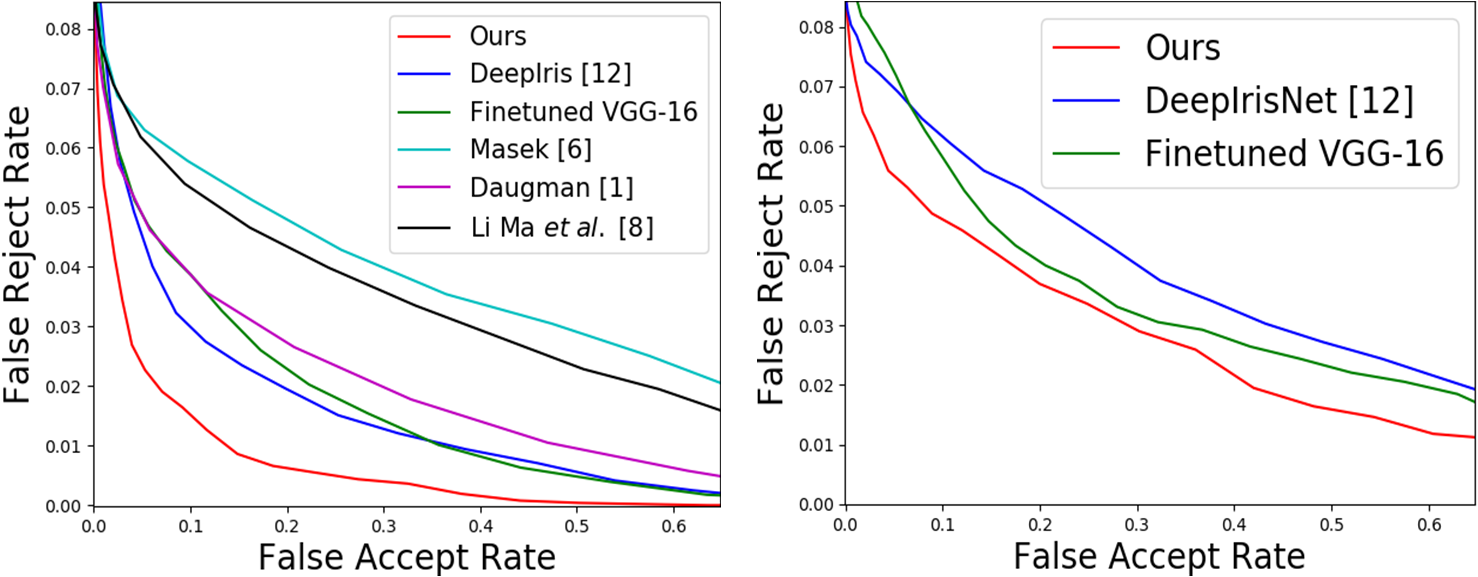}
    \vspace{-3mm}
    \caption{\scriptsize DET curve of: \textbf{Left:} comparing traditional and deep learning methods on CASIA.v4-Distance (Within Dataset), \textbf{Right:} comparing deep learning methods on CASIA.v4-Thousand (Cross Dataset)}
    \label{fig:within_cross_DB}
    \vspace{-6mm}
\end{figure}

 Only test set (of within and cross dataset configuration) of both the dataset is used for reporting iris verification performance. \\
\textbf{(a.) Within Dataset:} First, we compare efficacy of our proposed $CombNet_{E_{\theta}}^{EAP + TEL}$  with three traditional baselines of  Daugman \cite{1262028}, Masek \cite{masek2003recognition} and Ma \textit{et al.} \cite{ma2003personal}. Across both the metrics reported in Table \ref{WithinDBcompare}, our proposed framework outperforms all the three  baselines by notable margins. Next, we compare with the recent deep learning frameworks. We initially compare against pre-trained (on Imagenet) VGG-16 fine tuned  on CASIA.v4-Distance dataset similar to the work done by \cite{minaee2016experimental, nguyen2017iris}. Normalised iris of $224\times224$ resolution is input to VGG-16 framework. Though fine-tuning a pre-trained (on Imagenet) VGG-16 performs better than the traditional methods, yet  $CombNet_{E_{\theta}}^{EAP + TEL}$ proves to be superior than it. This can be primarily attributed to the fact that the kernels of VGG-16 were trained to learn structure and shape cues present in natural images and not texture-rich contents as prevalent in iris images. Thus, naively applying transfer learning across such disparate domains is sub-optimal. From Table \ref{WithinDBcompare}, we also observe that our proposed shallow $CombNet_{E_{\theta}}^{EAP+ TEL}$ performs better than DeepIrisNet \cite{gangwar2016deepirisnet}. This boost is primarily because of our systematic design choices. As argued before, our stage-wise training compels the network to focus more on discriminating iris textures which is further improved with incorporation of \textit{EAP} and \textit{TEL} layers. Also, for a iris dataset having paucity of annotated labels, it is more prudent to have less complex (parameter counts) models over deeper counterparts. Both DeepIrisNet as well as fine-tuned VGG-16 have much deeper and complex architectures for limited annotated iris datasets, and thus our model consistently outperforms those. Figure \ref{fig:within_cross_DB} depicts the DET curve of all the competing models of this phase. \\
\begin{table}[]
\centering
\vspace{-3mm}
\scriptsize
\caption{\scriptsize Comparison on CASIA.v4-Thousand (cross dataset configuration).}
\label{crossdb_comparison}
\begin{tabular}{|c|c|c|}
\hline
\textbf{Model} & \textbf{\begin{tabular}[c]{@{}c@{}}EER\\ (in \%)\end{tabular}} & \textbf{AUC} \\ \hline
\multicolumn{3}{|c|}{\textit{\textbf{CASIA.v4-Thousand}}} \\ \hline
DeepIrisNet & 6.6 & 0.033 \\
VGG-16 & 6.6 & 0.028 \\
\textbf{$CombNet_{E_{\theta}}^{EAP+TEL}$ (Proposed)} & \textbf{5.3} & \textbf{0.018} \\ \hline
\end{tabular}
\vspace{-1mm}
\end{table}
\textbf{(b.) Cross Dataset:} From Table \ref{crossdb_comparison}, it is evident that even in such challenging scenario, our proposed framework performs better than the comparing deep networks. This proves better generalization capability of our proposed framework over other deep learning frameworks. Figure \ref{fig:within_cross_DB} depicts the DET curve of one of the randomly selected folds of the competing deep nets. For fairness, same fold is chosen for all the comparing models. 

\noindent
\textbf{Reduction of Parameters:} There is an increased demand to run biometrics systems on mobile devices. So lightweight models are favored for inference. In Table \ref{ablation}, we compare number of parameters of our different architectural choices. We see that replacing full-connected layers of $CombNet_{E_{\theta}}$ with \textit{TEL} layer in $CombNet_{E_{\theta}}^{EAP + TEL}$ results in 46.72$\times$ reduction in parameters. From Table \ref{WithinDBcompare}, it can be observed that compared to VGG-16 and DeepIrisNet \cite{gangwar2016deepirisnet}, our model, $CombNet_{E_{\theta}}^{EAP + TEL}$ is respectively 46.62$\times$ and 100.41$\times$ cheaper in terms of parameters; yet our performance is better than those. It is suggested in this section to note that input to VGG-16 are normalised iris of dimension $224\times224$, while all other models have input iris images dimension of $512\times64$.
\vspace{-4mm}
\section{Conclusion}
\label{sec:conclusion}
\vspace{-2.4mm}
This paper proposes stage-wise texture aware training strategies for building reliable iris verification system under limited annotated data. This paper showcases benefits of unsupervised
auto-encoder based pre-traning as a good weight initializer for training networks with less data. Further, proposed \textit{EAP} and \textit{TEL} layers are shown to leverage local texture patterns of iris images. Our final framework is significantly lightweight and consistently outperforms competing baselines for within and cross dataset evaluations. Motivated by the success of auto-encoder based pre-training, in future, we wish to study the benefits of other recent generative models.

\bibliographystyle{IEEEbib}
\bibliography{reference}

\end{document}


%
\maketitle
\section{Architecture of Stagewise Training}
\subsection{Stage-1 Network Architecture}
To design the architecture of the encoder $\left(E_{\theta}\left(\cdot\right)\right)$ and the decoder  $\left(D_{\psi}\left(\cdot\right)\right)$ in Section 3.1.1 of the paper, we use 4 Convolutional Layers with Batch Normalization and Pooling  for $E_{\theta}\left(\cdot\right)$ and 4 Pixel Shuffle Layers~\cite{} for $D_{\psi}\left(\cdot\right)$. The details of the architecture are shown in Table~\ref{encoderDec}.
\begin{table}[h!]
\centering
\vspace{-5mm}
\caption{Configurations of various layers of $E_{\theta}(\cdot)$ and  $D_{\phi}(\cdot)$}
\label{encoderDec}
\begin{tabular}{lcccc}
\hline
\textbf{Type} & \textbf{Kernel} & \textbf{Stride} & \textbf{Padding} & \textbf{Output} \\
&&&&\textbf{Channels}\\\hline
\multicolumn{5}{c}{\textbf{Encoder}} \\ \hline
Conv & $5\times5$ & 1 & 2 & 32 \\
Batch Norm &  &  &  & 32 \\
Pooling & $2\times2$ & 2 & 0 & 32 \\
Conv & $3\times3$ & 1 & 1 & 64 \\
Batch Norm &  &  &  & 64 \\
Pooling & $2\times2$ & 2 & 0 & 64 \\
Conv & $3\times3$ & 1 & 1 & 128 \\
Batch Norm &  &  &  & 128 \\
Pooling & $2\times2$ & 2 & 0 & 128 \\
Conv & $3\times3$ & 1 & 1 & 256 \\
Batch Norm &  &  &  & 256 \\
Pooling & $2\times2$ & 2 & 0 & 256 \\
\hline
\multicolumn{5}{c}{\textbf{Decoder}} \\ \hline
Pixel Shuffle &  &  &  & 64 \\
Pixel Shuffle &  &  &  & 16 \\
Pixel Shuffle &  &  &  & 4 \\
Pixel Shuffle &  &  &  & 1 \\ \hline
\end{tabular}
\end{table}


\subsection{Stage-2 Network Architecture}
Our final Stage-2 network architecture, $CombNet_{E_{_\theta}}^{EAP+TEL}$ comprises of the encoder, $E_{\theta}\left(\cdot\right)$ of the previous section along with the \textit{TEL} layer as discussed in Section 3.1.3. The \textit{TEL} layer is used as the representative signature layer for the matching framework in the paper. The output of this layer is fed to a fully connected layer, culminating to the number of classes used while training. The details about different layers of the network are given in Table \ref{combnet}
\begin{table}[t]
\centering
\vspace{-7mm}
\caption{Configurations of various layers of {$CombNet_{E_{_\theta}}^{EAP+TEL}$}}
\label{combnet}
\begin{tabular}{lcccc}
\hline
\textbf{Type} & \textbf{Kernel} & \textbf{Stride} & \textbf{Padding} & \textbf{Output} \\
&&&&\textbf{Channels}\\\hline
Conv & $5\times5$ & 1 & 2 & 32 \\
Batch Norm &  &  &  & 32 \\
Pooling & $2\times2$ & 2 & 0 & 32 \\
Conv & $3\times3$ & 1 & 1 & 64 \\
Batch Norm &  &  &  & 64 \\
Pooling & $2\times2$ & 2 & 0 & 64 \\
Conv & $3\times3$ & 1 & 1 & 128 \\
Batch Norm &  &  &  & 128 \\
Pooling & $2\times2$ & 2 & 0 & 128 \\
Conv & $3\times3$ & 1 & 1 & 256 \\
Batch Norm &  &  &  & 256 \\
Pooling & $2\times2$ & 2 & 0 & 256 \\
Conv & $3\times3$ & 1 & 1 & 1024 \\
Batch Norm &  &  &  & 1024 \\ 
Fully Connected & & & & \#Classes\\
\hline
\end{tabular}
\end{table}
